\documentclass{article}

\usepackage{PRIMEarxiv}

\usepackage[utf8]{inputenc} 
\usepackage[T1]{fontenc}    
\usepackage{hyperref}       
\usepackage{xurl}           
\usepackage{booktabs}       
\usepackage{amsfonts}       
\usepackage{nicefrac}       
\usepackage{microtype}      
\usepackage{fancyhdr}       
\usepackage{multirow}
\usepackage{orcidlink}
\usepackage{graphicx}       
\graphicspath{{figures/}}     

\title{Towards Cross-Disaster Building Damage Assessment with Graph Convolutional Networks
\thanks{\textit{\underline{Citation}}: 
\textbf{Authors. Title. Pages.... DOI:000000/11111.}} 
}

\author{
  Ali Ismail\,\orcidlink{0000-0001-8636-1373} \\
  Department of Electrical and Computer Engineering \\
  American University of Beirut \\
  Beirut, Lebanon\\
  \texttt{ali.ai@live.com} \\
   \And
  Mariette Awad\,\orcidlink{0000-0002-4815-6894} \\
  Department of Electrical and Computer Engineering \\
  American University of Beirut \\
  Beirut, Lebanon\\
  \texttt{mariette.awad@aub.edu.lb} \\
}

\begin{document}
\maketitle

\begin{abstract}
In the aftermath of disasters, building damage maps are obtained using change detection to plan rescue operations. Current convolutional neural network approaches do not consider the similarities between neighboring buildings for predicting the damage. We present a novel graph-based building damage detection solution to capture these relationships. Our proposed model architecture learns from both local and neighborhood features to predict building damage. Specifically, we adopt the sample and aggregate graph convolution strategy to learn aggregation functions that generalize to unseen graphs which is essential for alleviating the time needed to obtain predictions for new disasters. Our experiments on the xBD dataset and comparisons with a classical convolutional neural network reveal that while our approach is handicapped by class imbalance, it presents a promising and distinct advantage when it comes to cross-disaster generalization.
\end{abstract}


\section{Introduction}
\label{sec:intro}
Change detection (CD) is a remote sensing technique used to detect changes between images taken at different times. It is widely used in many applications such as environmental \cite{song_change_2019} and urban monitoring \cite{zhang_detecting_2019}. Notably, CD has been instrumental in detecting building damage in disasters' aftermath \cite{robila_use_2006,gupta_xbd_2019}
Recent CD solutions have heavily relied on convolutional neural networks (CNN) \cite{khelifi_deep_2020} including those aimed at building damage detection \cite{wheeler_deep_2020}. Disaster-induced damage is a spatially spreading process therefore buildings in proximity sustain damage similarly. However, this aspect is ignored by CNNs which only consider the window defined by the convolution kernel. Graph convolutional networks (GCN) are an emerging architecture that generalizes stationary convolutions by taking into account non-euclidean relationships defined by non-local graph connections \cite{kipf_semi-supervised_2017}. This has lead to an improvement in many computer vision applications \cite{mou_nonlocal_2020,hong_graph_2020} including urban change detection \cite{saha_semisupervised_2020}.

Additionally, effective planning of rescue operations requires damage mappings to be obtained in a short time to minimize lives lost. However, data labeling and machine learning training can take weeks. This has been mainly addressed by pretrained generalizable models to infer on new disasters \cite{xu_building_2019,benson_assessing_2020,yang_transferability_2021}. While GCNs initially were unable to be used on unseen contexts\cite{kipf_semi-supervised_2017}, the SAGE (sample and aggregate) framework has been proposed to learn inductively on mini-batches of graphs and can be used to predict on unseen graphs \cite{hamilton_inductive_2018}.

To address the gaps highlighted in the field, we approach building damage CD as a node classification problem. Our graph construction uses building image representations as node features and a geometric triangulation to connect each building to its neighbors as opposed to the state-of-the-art which uses summary superpixel features for nodes and connects only adjacent superpixels \cite{saha_semisupervised_2020}. To leverage local and neighborhood building features, we present a novel hybrid architecture composed of a CNN backbone and a SAGE network. We test the cross-domain generalization of this model using an experimental design similar to \cite{xu_building_2019}. However, our experiments consider different types of disasters across multiple regions as opposed to earthquakes occurring in different locations. We use the xBD dataset for training and validation \cite{gupta_xbd_2019}.

The contributions of this paper are: 1) A novel graph node classification approach to disaster building damage CD, 2) A novel model architecture combining a CNN and a SAGE GCN to capture both local image features as well as neighborhood building similarities, 3) A cross-domain generalization of the proposed approach.

To the best of our knowledge, there is no prior work that studies the transferability of Graph SAGE models for building damage CD.

The rest of this document is organized as follows. In Section \ref{sec:methodology}, we explain our methodology. Section \ref{sec:experiments} details the experiments performed, and result interpretation while Section \ref{sec:conclusion} concludes the work with future research directions.

\section{Methodology}
\label{sec:methodology}
\subsection{Graph Data Formulation}
We create an undirected acyclic graph where each building is connected to its neighbors to exploit similarities between buildings in proximity. Each node represents one building defined by a rectangular envelope enclosing the building polygon. The node features are the concatenation of the pre and post image crops within the envelope. To unify crop size and reduce the memory footprint, we resize them to a width and height of $128 \times 128$. The edges are built using a Delaunay triangulation \cite{lee_two_1980} based on the building envelope centroid pixel coordinates. Edge weights are calculated according the similarity measure used by \cite{saha_semisupervised_2020} to build their adjacency matrix.

\begin{figure}[htb]
    \centering
    \includegraphics[scale=0.5]{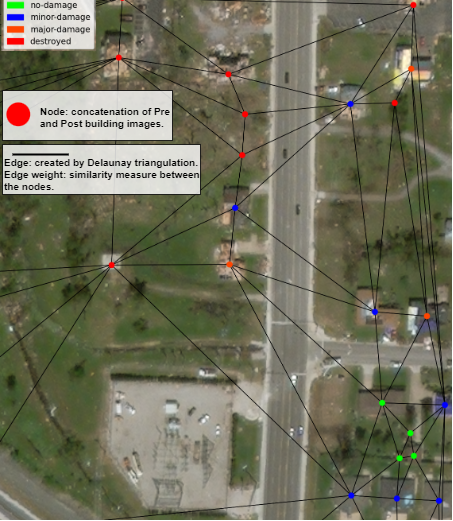}
    \caption{Subgraph implementation.}
    \label{fig:graph_form}
\end{figure}

In the SAGE paradigm, subgraphs are sampled from the large graph to create mini-batches. Since our graph is based on geographical proximity, every region can be represented by a single graph. However, storing this large graph to sample subgraphs proved to be too memory consuming due to the large number of buildings and the high dimensionality of the node features. The xBD dataset is divided into image chips each covering a certain geographical area and containing a non uniform number of buildings. Therefore, we use this predefined organization to build a subgraph out of each chip. Additionally, image chips with only one building and image chips with only one annotated building or with no labeled buildings were discarded. Figure \ref{fig:graph_form} shows a a view of an implemented subgraph.

\subsection{Model Architecture}
\begin{figure}[htb]
    \centering
    \includegraphics[width=\textwidth]{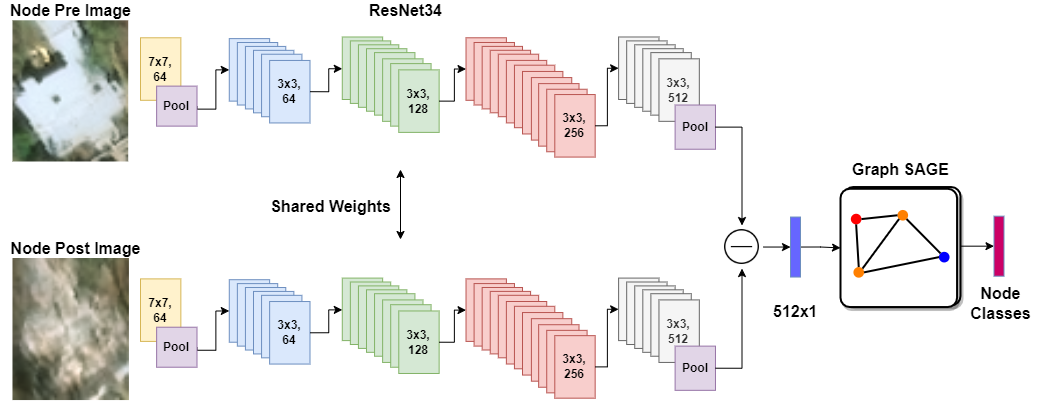}
    \caption{Architecture of the proposed model.}
    \label{fig:model_arch}
\end{figure}

We present a novel architecture based on a GCN with a CNN backbone which is necessary to reliably extract local image features while the downstream GCN is used to aggregate them with CNN features of connected nodes. The backbone is a Siamese ResNet34 network with the classification layers removed \cite{he2015deep} initialized with Imagenet weights. For the graph convolutions, we use the SAGE operator \cite{hamilton_inductive_2018}. The feature maps produced by the two ResNet streams are subtracted and inputted to the SAGE network whose output is the damage classification. All layers use the ReLu nonlinearity except for the final output layer which uses a Softmax activation. Figure \ref{fig:model_arch} shows the layout of the proposed architecture.

\section{Experiments and Results}
\label{sec:experiments}

\subsection{Experimental Setup}
The experiments were implemented using Python 3.8, Pytorch 1.7.1 and Pytorch Geometric 1.7.0 \cite{fey_fast_2019}. The code\footnote{\url{https://gitlab.com/awadailab/sage-project}} was executed on a virtual machine running on 8 cores of an AMD EPYC 7551 32-Core Processor with an Nvidia V100 32GB GPU. The Adam optimizer \cite{kingma2017adam} was used with the categorical cross entropy loss which included class weights to improve performance on minority classes \cite{wheeler_deep_2020}. Additionally, we merged the "major damage" and "destroyed" classes into a single class because of their ambiguity which would have degraded model performance as done in \cite{xu_building_2019}. The number of neurons, number of layers, dropout rate, learning rate and batch size were tuned empirically based on the performance on the test set. The resulting hyperparameters are two layers with 32 neurons and a dropout rate of 0.5. The batch size was set to 256 and the learning rate to 0.0003. For evaluation, we compute the accuracy, macro-averaged F1, weighted F1 and AUC scores.

\subsection{Experimental Procedures and Results}
At first, we train and test on the same disaster type (fire) and we use two fire disasters to make the training set sufficiently large. The following experiments test transfer from one disaster (flooding) and from multiple disasters (flooding, tornado and volcano) to a different disaster (fire). Lastly, we explore leaking a small number of test samples. As a benchmark, we adopt a traditional Siamese CNN identical to the Graph SAGE architecture except that the SAGE layers are replaced with fully connected layers. This comparison, shown in Table \ref{tab:sage}, allows us to isolate the contribution of the graph convolutions. 

\begin{table}[htb]
\centering
\caption{Comparison between the Graph SAGE and the Siamese CNN.}
\label{tab:sage}
\resizebox{\textwidth}{!}{%
\begin{tabular}{c|c|ccccc|ccccc}
\textbf{Train} & \textbf{Test/Hold} & \multicolumn{5}{c|}{\textbf{Siamese CNN}} & \multicolumn{5}{c}{\textbf{Graph SAGE}} \\ \hline
\multirow{4}{*}{\textbf{Socal + Portugal fire}} & \multirow{4}{*}{\textbf{Socal}} &  & \textbf{Acc} & \textbf{Macro F1} & \textbf{Weighted F1} & \textbf{AUC} &  & \textbf{Acc} & \textbf{Macro F1} & \textbf{Weighted F1} & \textbf{AUC} \\
 &  & \textbf{Train} & 0.8746 & \textit{\textbf{0.5994}} & 0.9036 & \textit{\textbf{0.9648}} & \textbf{Train} & \textit{\textbf{0.9110}} & 0.5642 & \textit{\textbf{0.9205}} & 0.9347 \\
 &  & \textbf{Test} & 0.8439 & \textit{\textbf{0.5267}} & 0.8696 & \textit{\textbf{0.8606}} & \textbf{Test} & \textit{\textbf{0.8672}} & 0.5042 & \textit{\textbf{0.8768}} & 0.8567 \\
 &  & \textbf{Hold} & 0.8581 & \textit{\textbf{0.5449}} & 0.8787 & \textit{\textbf{0.8609}} & \textbf{Hold} & \textit{\textbf{0.8870}} & 0.5315 & \textit{\textbf{0.8902}} & 0.8374 \\ \hline
\multirow{4}{*}{\textbf{Nepal flooding}} & \multirow{4}{*}{\textbf{Socal}} &  & \textbf{Acc} & \textbf{Macro F1} & \textbf{Weighted F1} & \textbf{AUC} &  & \textbf{Acc} & \textbf{Macro F1} & \textbf{Weighted F1} & \textbf{AUC} \\
 &  & \textbf{Train} & \textit{\textbf{0.6921}} & \textit{\textbf{0.6342}} & \textit{\textbf{0.7275}} & \textit{\textbf{0.8764}} & \textbf{Train} & 0.5092 & 0.3421 & 0.5487 & 0.5314 \\
 &  & \textbf{Test} & 0.4609 & 0.3223 & 0.5911 & \textit{\textbf{0.6229}} & \textbf{Test} & \textit{\textbf{0.5697}} & \textit{\textbf{0.3245}} & \textit{\textbf{0.6525}} & 0.6121 \\
 &  & \textbf{Hold} & 0.4356 & 0.315 & 0.5708 & \textit{\textbf{0.6601}} & \textbf{Hold} & \textit{\textbf{0.6075}} & \textit{\textbf{0.3275}} & \textit{\textbf{0.6727}} & 0.4999 \\ \hline
\multirow{4}{*}{\textbf{\begin{tabular}[c]{@{}c@{}}Nepal flooding + Joplin tornado\\ +   Puna volcano\end{tabular}}} & \multirow{4}{*}{\textbf{Socal}} &  & \textbf{Acc} & \textbf{Macro F1} & \textbf{Weighted F1} & \textbf{AUC} &  & \textbf{Acc} & \textbf{Macro F1} & \textbf{Weighted F1} & \textbf{AUC} \\
 &  & \textbf{Train} & \textit{\textbf{0.9283}} & \textit{\textbf{0.9008}} & \textit{\textbf{0.9306}} & \textit{\textbf{0.9877}} & \textbf{Train} & 0.7042 & 0.2824 & 0.5870 & 0.5199 \\
 &  & \textbf{Test} & 0.599 & \textit{\textbf{0.3832}} & 0.6928 & \textit{\textbf{0.7265}} & \textbf{Test} & \textit{\textbf{0.8744}} & 0.3110 & \textit{\textbf{0.8157}} & 0.6319 \\
 &  & \textbf{Hold} & 0.6082 & \textit{\textbf{0.3913}} & 0.6966 & \textit{\textbf{0.6597}} & \textbf{Hold} & \textit{\textbf{0.8542}} & 0.3072 & \textit{\textbf{0.7873}} & 0.6445 \\ \hline
\multirow{4}{*}{\textbf{\begin{tabular}[c]{@{}c@{}}Nepal   flooding + Joplin tornado\\ + Puna volcano + 10\% Socal Fire\end{tabular}}} & \multirow{4}{*}{\textbf{Socal}} &  & \textbf{Acc} & \textbf{Macro F1} & \textbf{Weighted F1} & \textbf{AUC} &  & \textbf{Acc} & \textbf{Macro F1} & \textbf{Weighted F1} & \textbf{AUC} \\
 &  & \textbf{Train} & \textit{\textbf{0.9666}} & \textit{\textbf{0.9522}} & \textit{\textbf{0.9671}} & \textit{\textbf{0.9963}} & \textbf{Train} & 0.7094 & 0.2873 & 0.5951 & 0.5394 \\
 &  & \textbf{Test} & 0.6659 & \textit{\textbf{0.4205}} & 0.7418 & \textit{\textbf{0.7173}} & \textbf{Test} & \textit{\textbf{0.8591}} & 0.3172 & \textit{\textbf{0.7959}} & 0.6765 \\
 &  & \textbf{Hold} & 0.6668 & \textit{\textbf{0.4308}} & 0.7405 & \textit{\textbf{0.6853}} & \textbf{Hold} & \textit{\textbf{0.8552}} & 0.3131 & \textit{\textbf{0.7899}} & 0.6665
\end{tabular}%
}
\end{table}

The Graph SAGE outperforms the Siamese CNN only on $20$ out of $48$ total occasions in terms of raw performance. The likely reason is that the Graph SAGE is being more severely degraded by the class imbalance. The Graph SAGE seems to outperform the Siamese CNN in terms of accuracy and weighted F1 score with an average increase of $14.74\%$ and $6.34\%$ on the hold set respectively. These two metrics ignore the impact of class imbalance. On the other hand, it falls behind most of the time in terms of macro-averaged F1 and underperforms systematically in terms of AUC with an average difference of $5.07\%$ and $5.44\%$ on the hold set respectively. Another possible reason is that given the geographical distribution of the samples, a large majority of subgraphs had no damaged buildings.

\begin{figure}[htb]
    \centering
    \includegraphics[width=\textwidth]{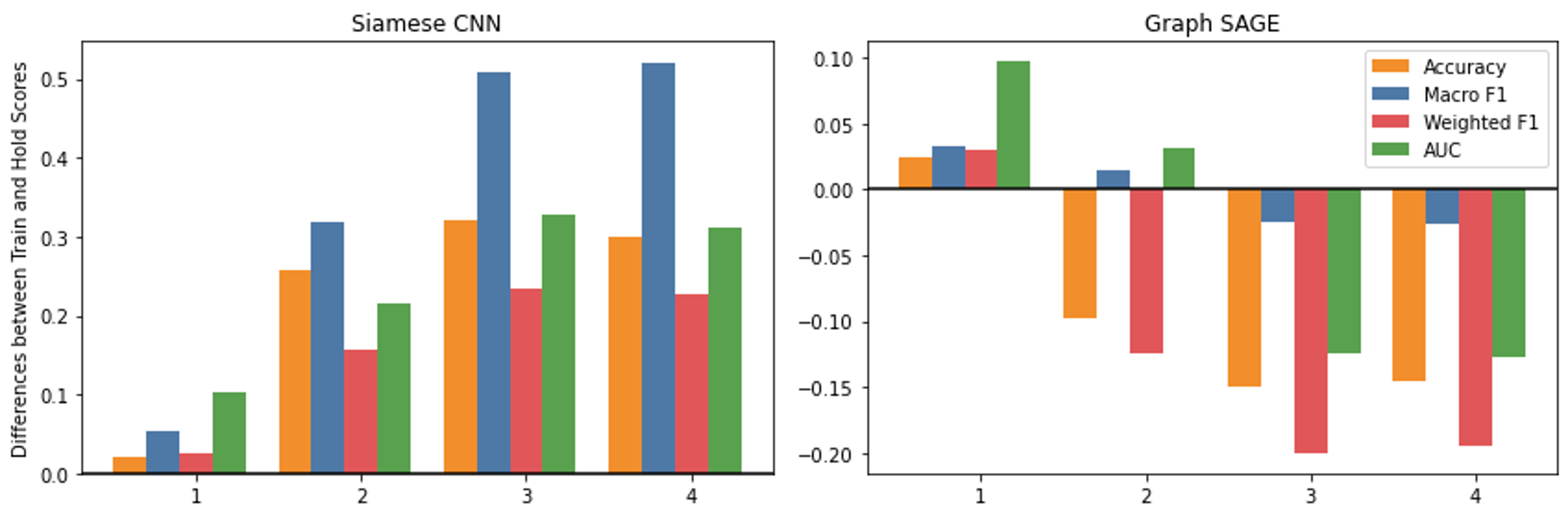}
    \caption{Difference between training and hold scores for the Siamese CNN and Graph SAGE. The x-axis indexes indicate the different experiments with different train - test configurations: 1) Fire - Fire, 2) Flooding - Fire, 3) Flooding + Tornado + Volcano - Fire and 4) Flooding + Tornado + Volcano + 10\% Fire - Fire.}
    \label{fig:sage_diff}
\end{figure}

In terms of cross-domain generalization, the Graph SAGE offers a more satisfactory behavior. We demonstrate this by computing the difference between the performance on the training set and the hold set (Figure \ref{fig:sage_diff}). For the first experiment, the Siamese CNN presents a lesser drop in performance. However, for subsequent experiments which are cross-domain, the difference for Graph SAGE is negative in most cases which means that performance on the hold set was higher than performance on the training set. Furthermore, the negative difference increases in magnitude as we move from left to right. Meaning that generalization improved when augmenting the training data with more diverse disasters. This implies that larger training data improves generalization to new domains. When the training data is supplemented with a small number of test samples, we achieve the best performance. On the other hand, the Siamese CNN shows a consistent positive generalization gap which is likely due to overfitting the training data. 

\section{Conclusion}
\label{sec:conclusion}
In the wake of disasters, building damage mappings provide essential intelligence for planning effective rescue and relief efforts. And the sooner this information is acquired, the more likely it is to rescue injured and trapped victims.

Like most geographical phenomena, disaster damage propagates such that buildings in proximity sustain damage in similar patterns. In this work, we formulated a graph-based building damage CD solution that captures both local patterns and building neighborhood relationships which are not captured by current approaches. Additionally, we focus on the cross-disaster generalization of our model which would enable our solution to be used to infer on new disasters without training thus accelerating the process. Although our results are not conclusive in terms of raw performance when compared to a classical CNN model, we note an advantage in generalizing to unseen and different disasters.

Based on the insights gained throughout this work, our future research includes devising more optimal subgraph sampling strategies and using graph-specific data balancing mechanisms to address the shortcomings of our approach.

\bibliographystyle{unsrt}  
\bibliography{references}

\end{document}